\documentclass{singlecolnew}

\usepackage{apacite}
\usepackage{cite}
\usepackage{verbatim}
\usepackage{multirow}
\usepackage[table]{xcolor}
\usepackage{stfloats,graphicx}
\usepackage{xcolor}
\usepackage[]{hyperref}
\hypersetup{pdftitle={DeepVeil: Deep learning for identification and face, gender, expression recognition under veiled conditions}, colorlinks=true, citecolor=blue, linkcolor=red}

\usepackage{algorithm}
\usepackage[noend]{algpseudocode}

\theoremstyle{TH}{

}

\theoremstyle{THrm}{

}

\theoremstyle{THhit}{

}

\makeatletter
\def\theequation{\arabic{equation}}

\def\BottomCatch{%
\vskip -10pt
\thispagestyle{empty}%
\begin{table}[b]%
\NINE\begin{tabular*}{\textwidth}{@{\extracolsep{\fill}}lcr@{}}%
\\[-12pt]
\end{tabular*}%
\vskip -30pt%
\end{table}%
} \makeatother

\begin{document}%

\setcounter{page}{1}

\LRH{Hassanat et~al.}

\RRH{Deep learning for identification and face, gender, expression recognition }

\VOL{x}

\ISSUE{x}

\PUBYEAR{2021}

\BottomCatch

\CLline

\title{\sf{\textbf{Deep learning for identification and face, gender, expression recognition under constraints}}}
\subtitle{}

\authorA{\sf{Ahmad B. Hassanat}}
\authorB{\sf{Abeer Albustanji}}
\authorC{\sf{Ahmad S. Tarawneh}}

\authorD{\sf{Malek Alrashidi}}
\authorE{\sf{Hani Alharbi}}
\authorF{\sf{Mohammed Alanazi}}
\authorG{\sf{Mansoor Alghamdi}}
\authorH{\sf{Ibrahim S. Alkhazi}}
\authorI{\sf{V. B. Surya Prasath}}
\affA{Faculty of Information Technology,\\
Mutah University,\\
Karak, Jordan\\
E-mail:hasanat@mutah.edu.jo}
\affB{Ministry of environment,\\
Amman, Jordan\\
E-mail: abeeralbustanji95@gmail.com}
\affC{Department of Algorithm and Their Applications,\\
E\"{o}tv\'{o}s Lor\'{a}nd University,\\
Budapest, Hungary\\
E-mail: ahmad.trwh@gmail.com}
\affD{Computer Science Department, Community College,\\ University of Tabuk,\\
Tabuk 71491, Saudi Arabia\\
E-mail: Mqalrashidi@ut.edu.sa}

\affE{Faculty of Computer and Information Systems, \\
Islamic University of Madinah,\\
Medina, Saudi Arabia\\
E-mail: hani.almoamari@iu.edu.sa}

\affF{Centre for Computational Engineering Sciences,\\ Cranfield University,\\ United Kingdom\\
Email: Alayda@ut.edu.sa}

\affG{Computer Science Department, Community College,\\ University of Tabuk,\\ Tabuk 71491, Saudi Arabia\\
Email: malghamdi@ut.edu.sa
}
\affH{College of Computers \& Information Technology,\\ University of Tabuk, \\Tabuk 71491, Saudi Arabia\\
Email: i.alkhazi@ut.edu.sa
}

\affI{Department of Electrical Engineering and Computer Science,\\
University of Cincinnati,\\
Cincinnati, OH 45267, USA\\
Email: prasatsa@uc.edu
}

\begin{abstract}
Biometric recognition based on the full face is an extensive research area. However, using only partially visible faces, such as in the case of veiled-persons, is a challenging task. Deep convolutional neural network (CNN) is used in this work to extract the features from veiled-person face images. We found that the sixth and the seventh fully connected layers, FC6 and FC7 respectively, in the structure of the VGG19 network provide robust features with each of these two layers containing 4096 features. The main objective of this work is to test the ability of deep learning based automated computer system to identify not only persons, but also to perform recognition of gender, age, and facial expressions such as eye smile. Our experimental results indicate that we obtain high accuracy for all the tasks. The best recorded accuracy values are up to 99.95\% for identifying persons, 99.9\% for gender recognition, 99.9\% for age recognition and 80.9\% for facial expression (eye smile) recognition.

\end{abstract}

\KEYWORD{Veiled-face recognition, deep learning, convolutional neural networks, age recognition, gender recognition, facial expression recognition, eye smile recognition.}

\maketitle
\section{Introduction}\label{sec:intro}

A wide variety of systems require reliable solid individual acknowledgment plans to either confirm or determine the identity of an individual requesting their services. The purpose of such schemes is to guarantee that the rendered services are accessed only by a legitimate user and no one else~\cite{jain2004}. By using biometrics, it is possible to establish an individual's identity or confirm based on "who she/he is", rather than "what she/he has" (e.g., a token, key) or "remembers" (e.g., PIN)~\cite{delac2004, hassanat2017victory,Hassanat2018, Shamaileh2019}. Systems based on what users remember and not based on what they own such as passwords are easy to penetrate by guessing or by a brute force dictionary attack and they are easy to be forgotten, unlike those that use biometrics that are hard to be lost or forgotten. Biometrics provide convenient authentications (as users are no longer required to remember complex passwords) while preserving a sufficiently high degree of security~\cite{jain2004}.

The use of biometrics is the statute of verifying a person's identity by imaging or measuring unique characteristics of that person. Typical phases of biometric include the acquisition of the data, so called enrollment, the feature extraction (of a template based on the data), the comparison and the storage. The biometric characteristic can be physiological such as a face, iris texture of the iris, fingerprints, hand geometry, DNA and fingerprints, it can behavioral, such as handwriting, keystroke, and way of walking~\cite{delbaere2014, hassanat2011visual}. Face recognition plays a major role in biometrics, the face is a global feature of human beings, and the method of acquiring face images is non-intrusive and can be captured at a distance~\cite{zhao2003}. Face recognition keeps being an active topic in computer vision research~\cite{ahonen2004}. There are several biometric systems but among the six known biometric attributes deemed by~\cite{heitmeyer2000} in a Machine Readable Travel Documents (MRTD) system facial features record the highest compatibility, such as security system, enrollment, machine requirements, renewal and public perception.

Face recognition is a function that human achieves effectively, even under unfavorable conditions like facial changes due to aging or poor lighting. This simple function of our brains has become a real challenge in the recent computer vision. Face recognition is of a functional importance in numerous applications, such as authentication in security systems such as bank machines or computers. Several methods to solve the problem of facial recognition have been studied. The greater part of them has concentrated on frontal face pictures, profile pictures, or geometric areas, (for example, of eyes, noses, and mouths)~\cite{sato1998,moreno2016robust,Tarawneh2019}.  Face recognition when the full face is visible is an extensive research area with accurate results. However, using only parts of faces (as in veiled-persons) is a challenging task. Partial face recognition (PFR) has turned into a rising issue with expanding requirements of identification from CCTV cameras and vision systems in cell phones, robots, and Smart home accessories. It is therefore important to recognize the arbitrary facial patch or face sample covered to enhance the intelligence of these systems~\cite{moreno2013robust,hu2013}. Despite the significance of the human face for social communication and person identification, some people in the population insist on veiling their faces for certain cultural and other purposes, such as soldiers on the battlefield, contagious disease (such as Coronavirus Disease - COVID19), and muslim women who get dressed Niqab (cloth cover the face except the eye area). Thus, there is a need for computer applications that can identify a veiled-face for many reasons for example: facilitating the identification process of muslim women, who get dressed Niqab, to identify them without asking them to reveal the whole face, in cases where some men wear Niqab for the purpose of hiding identity when doing criminal activities, to identify veiled-terrorists and other security reasons.

\subsection{Gender Recognition}\label{ssec:introgend}

Automatic gender recognition is now linked to the extent to which it is used in many programs and devices, especially because of the growth of social networking sites on the Internet and social media~\cite{dhomne2018}. Gender Recognition was begun with the problem in psychophysical studies to gender classification of the human face; it focuses on the types of perception in perceiving human visual processing and recognizing related features that can be used to distinguish between female and male individuals. Some investigations have demonstrated that the inconsistency between a female face and male face can be utilized viably to improvise the result of face recognition software in biometrics devices~\cite{dhomne2018}. Gender classification has always been an active area of research in the field of computer vision and artificial intelligence. The methodologies used to classify gender are extensively walk-based, body-based and face-based. The vast majority of the researcher's center on gender classification using faces images. But advances in computer vision and machine learning made it possible to do the same using whole body images and the images containing only partial information. Some of the important problems that researchers are still facing are facial differences such as blockage, expression changes, lighting changes, and the high-dimensional feature space~\cite{liew2016}. The traditional approach utilized in face-based gender recognition usually involves Multiple stages starting from image acquisition and processing and then reducing the dimensions and then extracting the features and finally classifying them. In addition to the need for prior knowledge in the field of application in order to determine the best advantage extracted. The performance of the recognition system is very dependent on the type of classifier chosen, which is in turn dependent on the feature extraction method used. It is hard to have a classifier that combines best with the selected feature extractor such that the best classification performance is accomplished. And any changes in the problem domain require redesigning the entire system~\cite{liew2016}.

\subsection{Age Recognition}\label{ssec:introager}

In recent years, many applications have been developed using face recognition, such as identification systems and security systems. But many researchers have proposed age-classification systems based on facial images. For more than 20 years, gender classification has been one of the most research topics in this area~\cite{ueki2006}. The classification of age using the face is one of the tasks in the vision of human and computer, which is the basis for many applications such as forensics or social media~\cite{rothe2018}. Human age is one of the most important personal traits which can be deduced directly through different patterns appear on the face that has recently been used in biometrics, cosmetology and security control, but it is still a difficult and interesting field. It is known that there are some general changes when people get older in age from childhood to adulthood such as the size of the face and the shape of the eye, nose, mouth, eyebrows, and lips~\cite{padme2014}. Age and gender play key roles in social interactions, but they still face some challenges, notably extreme blur (low-resolution), expressions, occlusions, etc~\cite{levi2015}.

\subsection{Facial Expression Recognition}\label{ssec:introfacialex}

Facial behavior is one of the most important signs of sensing the human emotions and intention of people. Based on recent developments in human-centered computing, the automated system for accurate and reliable facial expressions has emerging applications such as remote education, interactive and entertainment games, intelligent transportation systems, etc.~\cite{liu2014}. Facial expression recognition (FER) has remained a challenging and interesting problem~\cite{mollahosseini2016}. Thus deep neural networks~\cite{lecun2015deep} have increasingly been supported to learn representations of discrimination for FER~\cite{li2018}. To the best of our knowledge, part of the face was not used to distinguish facial expressions using traditional or modern methods such as deep learning~\cite{hassanat2017victory,Hassanat2018, Shamaileh2019, Tarawneh2019b, Tarawneh2018}.

During the collection of veiled-persons image database (VPI), we had trouble taking pictures of veiled-persons for personal reasons, especially males, many of whom refused to wear veils. However, we were able to collect pictures of $150$ people of both genders with $14$ pictures each, under different circumstances such as lighting, with a view to realism during taking pictures. We used this new dataset for testing facial expression recognition which is more challenging than a straightforward full frontal face views considered by other methods of the past.

\subsection{Motivation and Contributions}\label{ssec:motiv}

Face recognition plays a major role in biometrics, however some people in the population prefer to veiling their faces for cultural and other purposes such as soldiers on the battlefield, contagious disease - Hepatitis and recent Coronavirus - COVID19 pandemic mask mandates~\cite{aseeri2020modelling}, and muslim women who get dressed in Niqab. Thus, there is a need for automatic computer assisted applications that can identify a veiled-face using machine learning models:
\begin{itemize}
	\item Facilitating the identification process of muslim women who get dressed Niqab, to identify them without asking them to reveal the whole face.
	\item In cases where persons attempt to hide their identity for potential criminal activities.
	\item To identify veiled-terrorists.
	\item Other security reasons.
\end{itemize}

In addition, our work has several contributions that tackle challenging face, gender, expression identification problems, and can be summarized as follows:
\begin{itemize}
    \item Creating a new veiled-persons image (VPI) database, containing veiled-face images of 150 persons (male and female) from different age groups in the range of 8 to 78 years. The total number of images is 2100.
    \item Besides identifying a veiled-person, several other application are tested using the new dataset including veiled-person gender recognition, veiled-person age recognition, and veiled-facial expression recognition just from the revealed part of the face.
    \item Unlike the traditional methods which used hand crafted-based features extraction methods to solve the aforementioned problems, we used a popular deep pretrained architecture, VGG, to extract high-level features from different layers of the VGG model.
    \item The performance of the different systems has been tested and compared using several machine learning algorithms, namely, k-nearest neighbors (kNN), random forest (RF), Na\"{i}ve Bayes (NB), BayesNet (BN), and artificial neural network (ANN).
    \item Each of the proposed systems is tested under different dimensionality reduction parentage ($99\%$, $97\%$, and $95\%$) using the principle component analysis algorithm.
\end{itemize}

\subsection{Work Structure}\label{ssec:workst}

The reminder of the work is organized as follows. In Section~\ref{sec:lit}, we review the most recent works in the field of deep learning for face recognition, including from partially covered face images. This includes a detailed analysis of how each method works and what results are achieved with a description of the data sets used in each method. In Section~\ref{sec:prop}, we propose a new method for veiled-face classification and recognition of gender, age and the facial expression. We detail how the new method works, its advantages and disadvantages, and how the accuracy of the veiled-face classification using deep learning model can be increased. The specific database, veiled-persons image (VPI), that we have created for the purpose of this research work is described in detail, including the number of images, people, naming system, and capturing camera specifications. Detailed experimental results are given in Section~\ref{sec:results} base on the method studied here including comparisons with related models. Finally, conclusion and future works are presented in Section~\ref{sec:conc}.

\section{Literature Review}\label{sec:lit}

Face recognition has become one of the most researched topics in computer vision and biometrics since the 1970s. More recently, traditional face recognition methods have been superseded by deep learning methods based on CNNs~\cite{trigueros2018}. Facial recognition has been developed significantly through the emergence of deep learning. Deep neural networks have recently achieved major success in object recognition because of their remarkable learning potential. Therefore, this has been an incentive in verifying its effectiveness in facial recognition, which many researchers have done~\cite{sun2015}.

Though face recognition systems have broad applications, they are vulnerable to attack. Thus, the presentation attack detection (PAD) method is required to improve the security of face recognition systems. To beat the restrictions of previously suggested PAD methods~\cite{nguyen2018} have proposed a new PAD method that used hybrid features of handcrafted and deep features extracted from the images by the visible-light camera sensor. Their proposed method uses the multilevel local binary pattern (MLBP) to extract skin features from facial images, and CNN to extract deep image features to distinguish between real and offensive facial images, which have stronger discrimination ability than the single image features. They also used the support vector machines (SVM) in order to classify the features of the image to the real and attack classes, the classification error was significantly reduced to $0.456\%$ on a CASIA database. \cite{he2018} proposed a model combining the fully convolutional network (FCN), with Sparse Representation Classification (SRC) to propose a new approach to partial face recognition, called Dynamic Feature Matching (DFM), to treat partial facial images regardless of size and without needed to align the face. Their proposed DFM method has achieved remarkable accuracy with high efficiency on various partial face databases, including LFW, YTF, and CASIA-NIR-Distance databases in comparison with state-of-the-art Partial Face Recognition (PFR) methods. \cite{farfade2015} have proposed a method based on deep learning, called Deep Dense Face Detector (DDFD), which is able to detect faces in a wide range of orientations using a single model. \cite{sun2017} have proposed a new method for face detection using deep learning to improve the state-of-the-art faster RCNN framework for generic object detection, by combining a number of strategies, including feature concatenation, hard negative mining, multi-scale training, model pre-training, and proper calibration of key parameters. Their proposed scheme outperformed some of the state-of-the-art face detection methods compared, making it the best model in terms of ROC curves comparing to the other methods tested on the FDDB benchmark. In addition,~\cite{sun2014} have proposed a method to learn a set of high-level feature representations through deep learning, referred to as Deep hidden IDentity features (DeepID), for face verification. The features are built on top of the feature extraction hierarchy of deep CNN and are summarized from multi-scale mid-level features. By representing a large number of different identities with a small number of hidden variables, highly compact and discriminative features are acquired. And therefore, achieved $97.45\%$ face verification accuracy. 

In a related work,~\cite{sun2015} proposed two deep neural networks architectures, where they reused the basic elements of GoogLeNet and VGG net for face recognition, the resulting network DeepID3 was rebuilt from inception layers and stacked convolution proposed in GoogLeNet and VGG net to make it appropriate for face recognition. Supervisory signs for identity verification were added to each of the intermediate and final feature extraction layers throughout the training. Their method achieved $99.53\%$ facial verification accuracy on the faces scanned in life wild database (LFW).
 
~\cite{hassanat2017} proposed a method  to verify the ability of a computer system to identify a veiled-person from the revealed part of the face alone, employing  some simple geometric and texture features, a new veiled-person image  database (VPI) was created for the purpose of their study using a mobile phone camera. The images were taken for $100$ different individuals (both male and female) from different age groups in the range of $13$ to $50$ years, over two sessions, The total number of images is $1200$, this method achieved $88.63\%$ to $97.22\%$ person identification accuracy. In their proposed system, two methods were implemented to extract distinct features from VPI, geometric features; such as edges, and texture features; such as mean, variance, skewness. Before extracting the features, the captured images were resized to $25\%$ of the original size with the aim of speeding the process.

\cite{hu2013} have introduced a new method to identify persons from their partial face images using the distance of an instance-to-class, which is based on representation of the local features. Their experiments were conducted on two sets of widely used data: LFW dataset and AR dataset. The method achieved up to $98\%$ accuracy. \cite{mahbub2016} proposed a face detection method based on part of the face. This method was designed to detect the partially cut and covered faces that were captured using a mobile phone  camera for continuous authentication. The performance of the face detector was evaluated on the Active Authentication dataset (AA-01), achieving high accuracy compared to some other methods.

~\cite{teo2007} used parts of frontal face images such as eye, nose, and mouth for personal authentication, the frontal human eye images were generated from Essex dataset with $153$ subjects, the partial face images were tested with non-negative matrix factorization (NMF), local NMF (LNMF) and spatially confined NMF (SFNMF), Their experimental results showed  that  the  LNMF performed better achieving $95.12\%$ recognition rate. \cite{akbar2015} proposed an arithmetical model for face recognition, where he investigated several feature extraction methods such as discrete wavelet transform (DWT), discrete sine transform (DST), local binary patterns (LBP) and local phase quantization (LPQ), with several classifiers such as SVM, probabilistic neural network (PNN) and k-nearest neighbors (kNN). The hybrid feature vector of DWT and DST achieved the best performance of $92.1\%$ accuracy using SVM. 

~\cite{ranjan2016} proposed a multi-task deep learning method called HyperFace for simultaneously detecting faces, localizing landmarks, estimating head pose and identifying gender using deep convolution neural networks (DCNNs). Their proposed method fuses the intermediate layers of a deep CNN using a separate CNN followed by a multi-task learning algorithm that operates on the fused features, this allowed the method to capture both global and local features of the targeted face image,  which achieved high results.

\cite{chen2018} proposed a face detection model named adversarial occlusion-aware face detector (AOFD) to address the issue of face occlusions. Their proposed model is able to find faces with a few exposed face features with extremely high confidence and maintain high accuracy for even detectable faces. A deep adversarial network was used in their proposed model to generate face samples with occlusions from mask generator. AOFD has achieved superior performance for face detection and masked face detection up to $97.88\%$ by training on benchmark dataset for general face detection such as FDDB.

In~\cite{huang2020curricularface}  proposed a new loss function, which is called Adaptive Curriculum Learning loss (CurricularFace). To help prioritizing the training of the hard examples, the proposed loss function claims to offer better training strategy for face recognition, as shown by the experiments conducted on common datasets.

In addition,~\cite{shi2020towards} proposed a face recognition framework called URFace. The proposed framework claimed to be able to recognise wide variety of faces under real-life conditions, including low resolution, occlusion and head pose. The proposed method rely on a new augmentation method, which works by using multiple sub-embeddings in order to make the training process smoother. The proposed framework recorded a state-of-the-art results on challenging datasets. More information can be found in a recent  survey of occluded and unoccluded Face Recognition~\cite{xu2021survey}.

\subsection{Gender Recognition}\label{ssec:gen}

DCNN has become a generic approach to gender recognition achieving many successes~\cite{dhomne2018} were the first to use the VGGNet to predict gender with celebrity face image database. One of the methods used in this work was the transfer learning, where a pre-trained DCNN achieved improved the performance up to $7\%$ and $4.5\%$ compared to the other gender recognition methods compared. In an earlier work~\cite{chi2006} were the first to use the CNN to develop a system for automatic gender recognition, which can detect faces in arbitrary size images, and then recognize gender. This system was tested on two different image databases, achieving $97.2\%$ recognition rate on the FERET database, and  $85.7\%$ recognition rate on images collected from the Web and the BioID face image database.

\cite{liew2016} proposed an optimized CNN architecture for real-time gender classification based on facial images. The number of processing layers was reduced in CNN to only four by fusing the convolutional and sub sampling layers,  cross-correlation was applied in the processing layers instead of convolution. The performance of their proposed CNN model was evaluated on two publicly available face databases, achieving $98.75\%$ recognition rate on SUMS, and $99.38\%$ on AT\&T.

\cite{hassanat2017} presented a model to measure the ability of the computer system to gender classification a veiled-person from the revealed part of the face alone depending on the Geometric features. Their proposed method achieved a success rate of $99.41\%$ for gender classification.

\subsection{Age Recognition}\label{ssec:ager}

\cite{ueki2006} proposed  a two-phased approach based on two-dimensional linear discriminant analysis (2DLDA) and LDA (2DLDA+LDA) to classify the age group using facial images under different lighting conditions. WIT-DB was created using images from about $5500$ different Japanese subjects (about $2500$ females and about $3000$ males) with $1$ to $14$ images of each subject. Most facial expressions in these images are normal except for smiles in some images. Their method achieved accuracy rates of $46.3\%$ for age groups within $5$ years, $67.8\%$ for age groups within $10$ years, and $78.1\%$ for age groups within $15$ years.              

\cite{levi2015} have proposed a simple convolutional network architecture, reducing the number of parameters and opportunities for processing and used when the amount of learning data is limited to estimate age and gender. Their proposed network was evaluated on the newly released audience face images, the evaluation results showed high age recognition rate compared to some other methods. The recent work of~\cite{rothe2018} used the CNNs of the VGG16 structure that was previously trained on ImageNet for image classification, their proposed form was named the Deep EXpectation (DEX), which relies on strong facial alignment to estimate real and apparent age. Their proposed method was evaluated on the IMDB-WIKI database, the largest public dataset to date with age and gender annotations, showing state-of-the-art results. \cite{gonzalez-briones2018} proposed a hybrid structure that included CNN, which aimed to extract features from input images; and extreme learning machine (ELM). The hybrid method is employed for gender and age classification. The proposed method was evaluated on  two popular face databases; namely, MORPHII and Adience benchmarks to classify inputs into four age groups, Children, Youth, Adults and Elderly, the best success rate achieved when using Fisher-faces with Gabor filter.

\subsection{Facial Expression Recognition}\label{ssec:fac}

\begin{table}
\centering
	\caption{Recognition tasks tackled by our proposed work compared to other related works.}\label{tab:Uniqueness}
	\begin{tabular}{llllll}
	\hline
	 Reference                   & Face       & Veiled-face & Age        & Gender     & Facial expressions \\ \hline
	Our proposed           & $\times$   & \checkmark  & \checkmark & \checkmark & \checkmark        \\
	\cite{hassanat2017}    & $\times$   & \checkmark  & $\times$   & \checkmark & $\times$          \\
	\cite{teo2007}         & \checkmark & $\times$    & $\times$   & $\times$   & $\times$          \\
	\cite{hu2013}          & \checkmark & $\times$    & $\times$   & $\times$   & $\times$          \\
	\cite{liew2016}        & $\times$   & $\times$    & $\times$   & \checkmark & $\times$          \\
	\cite{sun2015}         & \checkmark & $\times$    & $\times$   & $\times$   & $\times$          \\
	\cite{lian2016}        & $\times$   & $\times$    & $\times$   & \checkmark & $\times$          \\
	\cite{yu2015}          & $\times$   & $\times$    & $\times$   & $\times$   & \checkmark        \\
	\cite{rothe2018}       & $\times$   & $\times$    & \checkmark & $\times$   & $\times$          \\
	\cite{dehshibi2010new} & $\times$   & $\times$    & \checkmark & $\times$   & $\times$          \\ \hline
	\end{tabular}
\end{table}

Facial expressions are not used only to express our feelings, but also to provide important communication signals during social interaction, such as our level of interest. It is reported that the facial expressions have a significant effect on the listener; about $55$ percent of the impact of spoken words depends on facial expressions and eye movements of the speaker~\cite{ghosh2015}. There are many obstacles make facial expressions difficult to be recognized, such as occlusion, various head poses, low image resolution, and lighting conditions~\cite{luo2016}. Having been a challenging task, facial expression recognition has attracted many researchers to work on the entire face, however, a few have worked  on part of the face, since partial facial expression recognition is more challenging task.

\cite{mollahosseini2016} have worked on the whole face; proposing a new deep neural network architecture to automatically identify facial expressions. Their proposed network consists of two convolutional layers, each followed by max pooling and then four Inception layers. The network is a single-component structure that takes the facial images as an input and classifies them to any of the six basic expressions (anger, disgust, fear, happiness, sadness, and surprise). Their proposed method was tested on seven facial expression databases, namely, DISFA, FERA, SFEW, FER2013, MultiPIE, MMI, and CK. The results of the proposed architecture were better than that of the traditional convolutional neural networks in terms of accuracy and training time. ~\cite{yu2015} proposed a DCNN based facial expression recognition method. The proposed method includes a face detection module assembled from three state-of-the-art face detectors, followed by a classification module assembled from multiple DCNN. Their proposed method was tested on the FER and SFEW data sets. The highest accuracy achieved was $61.29\%$ on the SFEW dataset. \cite{Kotsia15} have worked on partial facial expression recognition, attempting to find  which part of the face (lower, upper, left or right) provides more discriminant information about each facial expression. They found that the lower part of the face (mouth region) provide more information about anger, fear, happiness and sadness, while information about disgust and surprise are well preserved by the upper part of the face (eyes region). In this work, where veiled-faces are used, all parts of the face are occluded, except for the eyes region, from which we attempt to recognize smiles, age and gender, in addition to identifying the person. This makes our work unique, more challenging and important. To the best of our knowledge, this is the first paper to work on veiled-faces solving the four aforementioned problems altogether in one paper, one exception is our previous work~\cite{hassanat2017}. However, our previous work was based on a smaller veiled-faces database, which does not contain age and smile information in addition to using the traditional hand-crafted features. While this work is based on the use of the deep features and a larger and wealthier veiled-faces database. Table~\ref{tab:Uniqueness} shows how the proposed work is compares from the existing works in terms of various recognition tasks.

\section{Design and Methodology of the Proposed Approach}\label{sec:prop}

\subsection{Veiled-Persons Identification (VPI-New) Database}\label{ssec:data}

\begin{table}
\centering
	\caption{Naming system used in our VPI-New database.}\label{tab:vpidatabase}
	\begin{tabular}{llllllll}
	\hline
	\#		&	Session	&	Person	&	Gender	&	Age	&	Image\#	&	facial exp.	&	File name\\
	\hline
	1	&	S1	&	P2		&	M	&	14	&	1	&	N	&	S1-P2-M-14-1-N\\
	2	&	S2	&	P100		&	F	&	36	&	2	&	S	&	S2-P100-F-36-2-S\\
	3	&	S2	&	P150		&	F	&	24	&	5	&	N	&	S2-P150-F-24-5-N\\
	4	&	S1	&	P57		&	F	&	55	&	2	&	S	&	S1-P57-F-55-2-S\\
	5	&	S2	&	P148		&	M	&	34	&	2	&	N	&	S2-P148-M-34-2-N\\
	\hline
	\end{tabular}
\end{table}

\begin{figure}[t]
\centering
	\includegraphics[width=8cm]{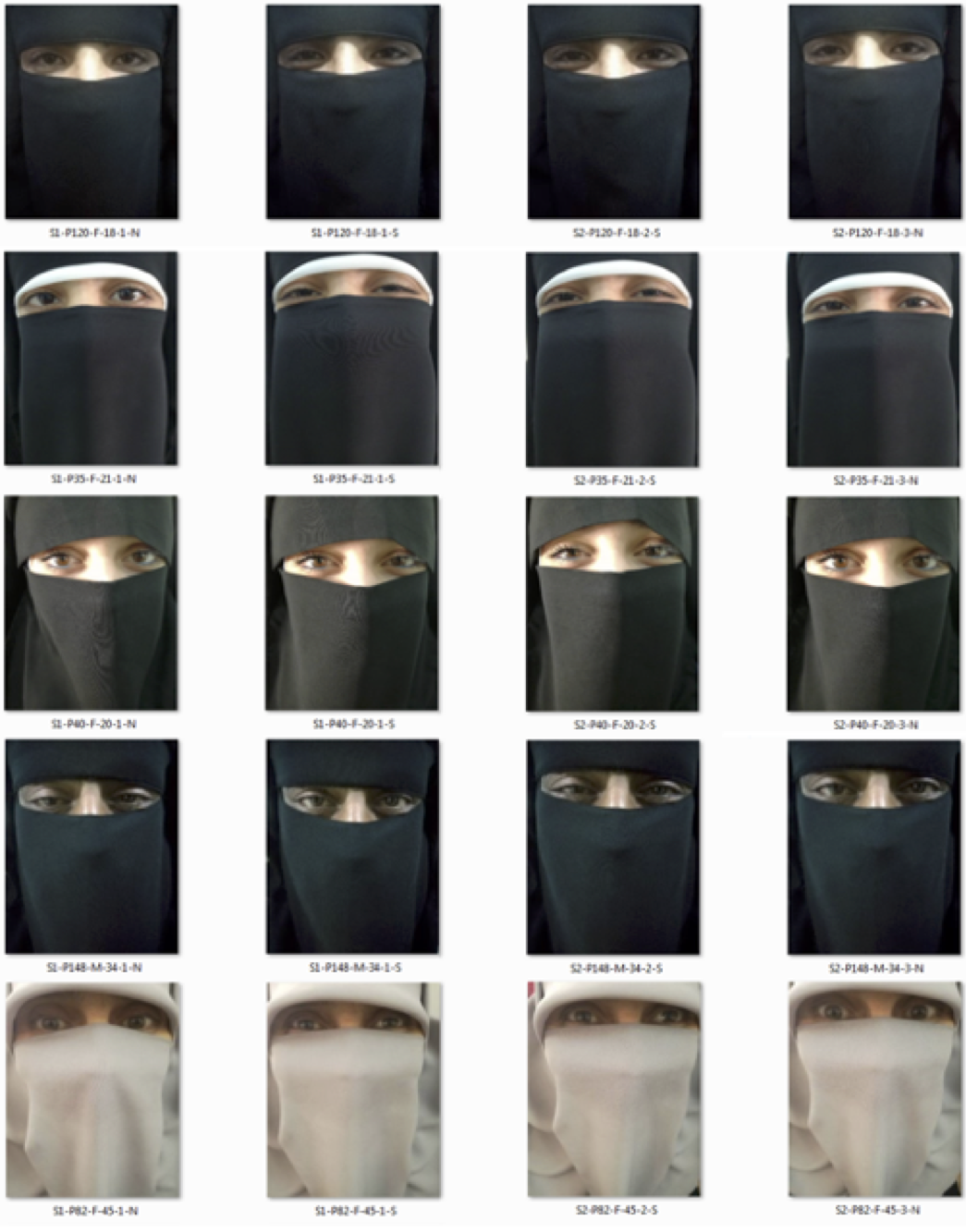}
	\caption{Sample images from VPI-New database with the same subject in each row, the colors of the veils were black and white, and the total number of images is $2100$ images. }\label{fig:vpidatabase}
\end{figure}	

The main objective of this work is to test the ability of an automated computer system to identify persons, their genders, their ages, and their facial expression, all from veiled-face images, employing one of the most common deep learning methods. Most of the standard database of face recognition consists of the whole face images, front or face profile, except for the VPI-Old database, which is created by~\cite{hassanat2017}. However, we found some limitations in this database, such as the relatively small number of images, which allows for ineffective training using deep learning, moreover, the images of the VPI-Old database does not contain facial expressions, such as eye-smile. Therefore, and for the purposes of this study, we have created a new database to satisfy the main objective of this paper. 

VPI-New is a new database created to identify persons, classify gender, age, and facial expressions (eye-smile) from veiled-face images. The images were taken while the subject was sitting on a chair, the camera was too close to the subject, the camera was moving a bit after each shot so that the images were taken from more than one angle and more than a degree of widening the eye. What  follows shows the design guidelines we followed while creating the VPI-New database:
\begin{itemize}
	\item[-] A HUAWEI P9 Lite mobile phone camera ($13$-megapixel, $3120 \times 4160$) was used to capture the veiled-face images of $150$ subjects ($41$ male and $109$ female) from different age groups in the range of $8$ to $78$ years. The images were taken in two different sessions (sessions 1  and 2) with seven pictures per session for the same subject. In each session, two images were taken in smiling mode and five images were taken in the normal mode. The total number of images of VPI-New database is $2100$.
	\item[-] This VPI-New is designed primarily for the purpose of this study; namely, identifying persons, distinguishing between male and female, age recognition, and facial expression; namely, eye-smile.  
	\item[-] The Images were taken in different offices and under uncontrolled lighting conditions (office environment).
	\item[-] Distances when taking pictures are not restricted, the distance between the camera and the veiled-person may vary in the range (from $30$ to $50$ cm).
	\item[-] All images were resized to $224 \times 224$ to fit the input layer in the VGG19 architecture before extracting features.
\end{itemize}

The file name starts with the session number (session 1 - S1 or session - S2), then (P) followed by the subject's number, then the subject gender (male - M or female - F), which is followed by the their age (in years), then the image number within each session ranging from 1 to 7, and finally facial expressions (Normal - N or Smile - S). Table~\ref{tab:vpidatabase} provides some examples of the naming system of the image files of our VPI-New database.

Figure~\ref{fig:vpidatabase} shows a sample of images from our VPI-New database. Note that each row shown in the figure is dedicated to a specific subject. These images were captured in different sessions and with both facial expressions (normal and eye-smile), under different conditions such as the difference in light, the widening of the eyes, the varied areas of the exposed part of the face, and type of Niqab used.

\subsection{Proposed Approach}\label{ssec:prop}

\begin{figure}
\centering
	\includegraphics[width=8cm]{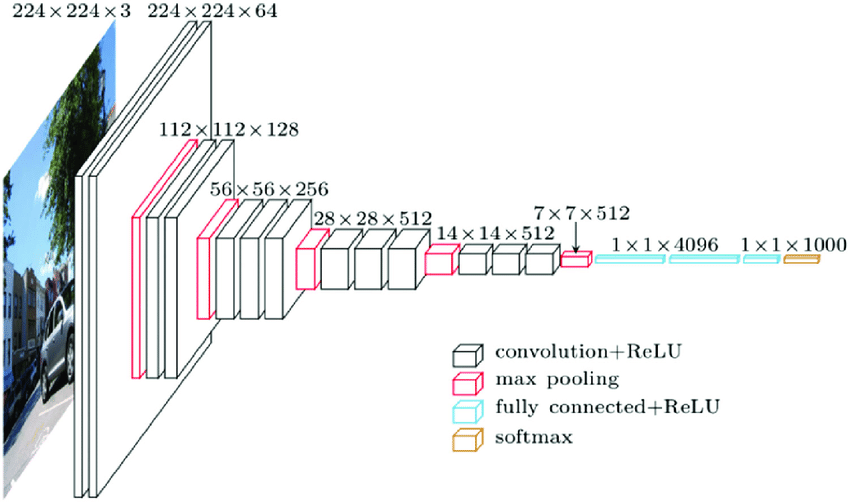}
	\caption{Typical architecture of the VGG model~\cite{simonyan2014very}. }\label{fig:vggarchi}
\end{figure}	

\begin{algorithm}\caption{FC6 and FC7 layer features extraction}\label{algo1}
	Input: Veiled-face image\\
	Output: Feature vectors of FC6 and FC7
	\begin{algorithmic}[1]
		\Procedure{Start}{}
		\State Read images from a folder of the VPI-New database
		\State Store the name of the files that represent the class 
		\State Load Pre-train VGG19
		\State Convert grey-scale images to RGB images
		\State Resize all images size to $224 \times 224$ to fit the input layer in VGG19
		\State Select which layer the features will be extracted from (FC6 or FC7)
		\State Extract features from FC6 layer or FC7 layer both separately
		\State Getting features vector for each layer
		\EndProcedure
	\end{algorithmic}
\end{algorithm}

\begin{figure}
\centering
	\includegraphics[width=0.9\textwidth]{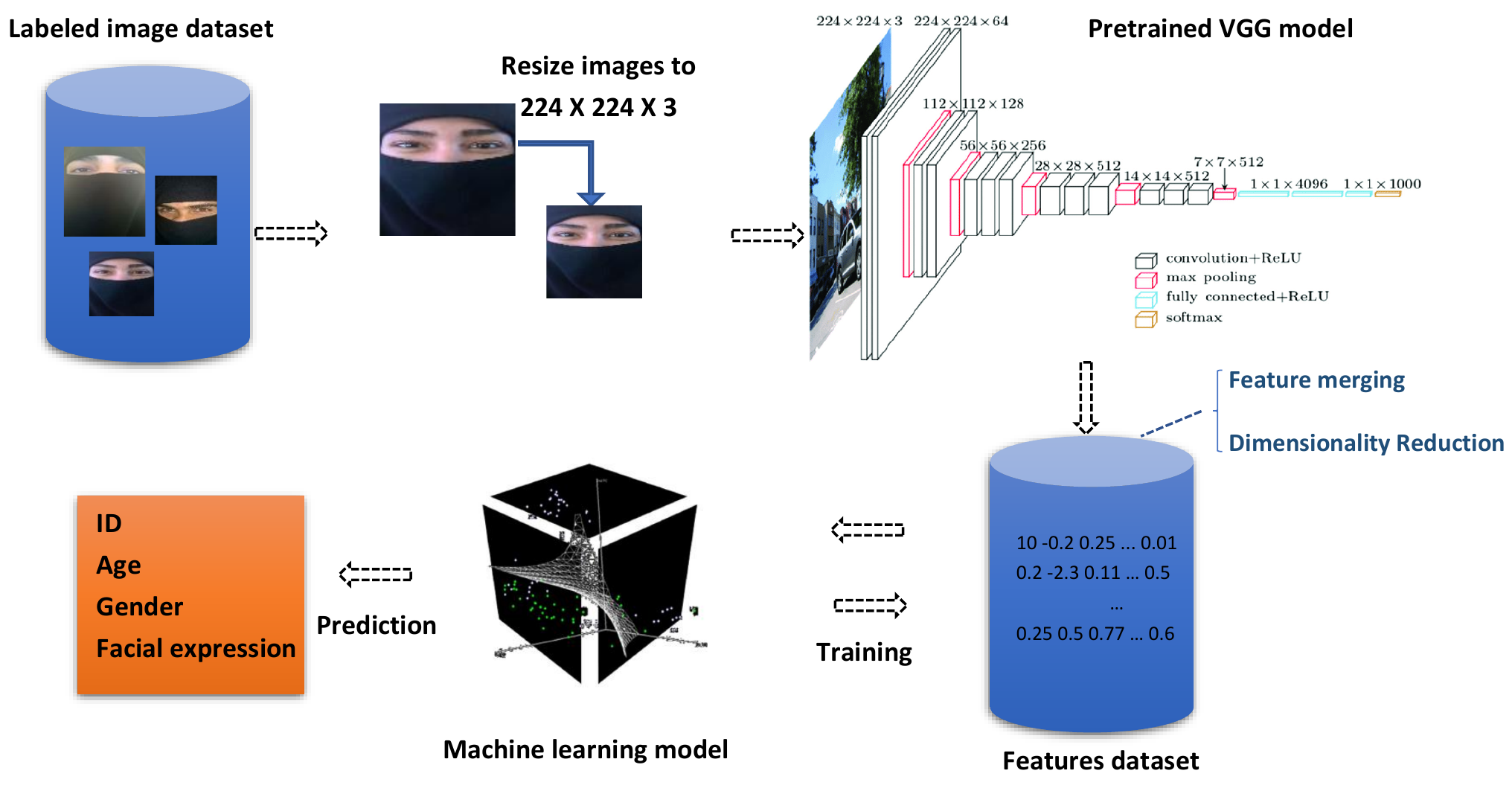}
	\caption{The flow diagram of the proposed framework }\label{fig:blockdiagram}
\end{figure}

Conversion of the input image into a set of features is called feature extraction. In computer vision, the feature extraction process is a special form of dimensionality reduction, so that is the input data will be converted into a reduced representation set of features (called feature vector).  The main goal of feature extraction is to obtain the most important information from the raw data and represent that information in a lower dimensionality space~\cite{kumar2014}. In order to extract features, we used Pre-trained VGG-19 (a typical architecture is presented in Figure~\ref{fig:vggarchi}), which was pre-trained on the ImageNet database~\cite{deng2009}. The obtained deep features were classified by some WEKA classifiers~\cite{witten2016}. 

\begin{table}
\centering
	\caption{The number of the deep features before and after applying the PCA.}\label{tab:nofeat}
	\begin{tabular}{l|l|llll}
	\hline
	&	&	Identify 	&	Gender 		&	Age 			&	facial expression \\
	&	&	persons	&	recognition	&	recognition	&	recognition\\
	\hline
	Raw 			&	FC6	&	4096		&	4096		&	4096		&	4096\\
	features		&	FC7	&	4096		&	4096		&	4096		&	4096\\
			\hline
	PCA 99\%		&	FC6	&	446		&	446		&	442		&	446\\
			&	FC7	&	157		&	157		&	157		&	157\\
			\hline
	PCA 97\%		&	FC6	&	208		&	208		&	208		&	208\\
			&	FC7	&	149		&	78		&	78		&	78\\
			\hline
	PCA 95\%		&	FC6	&	137		&	137		&	137		&	137\\
			&	FC7	&	56		&	56		&	56		&	56\\
	\hline
	\end{tabular}
\end{table}

It is worth noting the following characters of the extraction and classification tasks considered in this paper:
\begin{itemize}
	\item For the purpose of extracting features and classifying them to identify persons from their veiled-face images, the database is divided into $150$ folders, each contains $14$ images of a different subject. 
	
	\item In the case of gender classification, the database is divided into two folders, each contains $41$ Male and $109$ Female subjects.

	\item In the case of age classification, the subjects were categorized into four categories - Children, Youth, Adults, and Elderly. Hence, the database is divided into two folders, each contains the images of a certain age category, similar to~\cite{gonzalez-briones2018}, however they are slightly different in the number of subjects belonging to the last two categories, this is due to the lack of elderly people who volunteered to give their images. The number of subjects of the first class (Children which represents the age groups below $18$ years) is $36$, the number of subjects of the second class (Youth which represents the age groups of $19$ to $30$ years) is $75$, the number of subjects of the third class (Adults which represents the age groups of $31$ to $50$ years) is $33$, and the number of subjects of the fourth class (Elderly which represents the age groups of $51$ to $77$ years) is $6$.
	
	\item  In the case of the classification of facial expressions, the files were divided into two files - Eye-Smile, and Normal. The number of images of the first class (Normal) is $1500$ images and the number of images of the second class (Eye-Smile) is $600$ images.
\end{itemize}

In order to obtain deep features, we used the fully connected layer (FC6) and fully connected layer (FC7) of the VGG19 model. Algorithm~\ref{algo1} illustrates the steps of extracting these deep features to identify persons, and recognize gender, age and facial expression. Figure~\ref{fig:blockdiagram} shows a visual diagram of the proposed framework. VGGNet is one of the most preferred choices for extracting features from images~\cite{simonyan2014very}. Typically, VGGNet consists of $16$ layers or $19$ layers. Since VGG19 is deeper, we have used it to extract our features after resizing the images to $224 \times 224$ to fit the input layer. 

The output of Algorithm~\ref{algo1} is two  feature vector for each image, obtained from each layer (FC6 and FC7), each contains $4096$ features.  The resultant feature vectors of all the database is used for training and testing.

Due to the large size of the extracted features ($2100 \times 4097$), it is recommended to use some kind of dimensionality reduction~\cite{Hassanat2018b, Tarawneh2019}. Here, we used the Principal Component Analysis (PCA) for dimensionality reduction with three different percentages of data variance, namely $99\%$, $97\%$ and $95\%$. Since we expect that the number of the resultant principal components will be reduced as the percentage of the data variance is decreased, however keeping less data might significantly affects the classification results. The resultant PCA features are used for the four classification tasks, and Table~\ref{tab:nofeat} shows the number of features obtained after applying the three percentages of data variance. Note that this is done on both layers FC6 and FC7.

\begin{algorithm}\caption{Min, Max, Mean Merge with FC6 and FC7 layer features}\label{alg2minmaxmean}
	Input: Feature vector of FC6 and Feature vector of FC7\\	
	Output: Three Final feature vectors, one from each method (Min, Max and Mean),  FFVmin, FFVmax and FFVmean.
	\begin{algorithmic}[1]
	\Procedure{Start}{}
	\State load feature vector of FC6 layer
	\State load feature vector of FC7 layer
	 
	\State FFVmin= Min(FC6, FC7) as shown in Equation~\ref{E:FFVmin}
	\State FFVmax= Max(FC6, FC7) as shown in Equation~\ref{E:FFVmax} 
	\State FFVmean= Mean (FC6, FC7) as shown in Equation~\ref{E:FFVmean} 

	\EndProcedure
	\end{algorithmic}
\end{algorithm}

Since there is no a priori knowledge about which layer (FC6 or FC7) being the most representative, we propose merging both layers. For merging we investigated 3 different methods: the arithmetic mean, minimum and maximum of the two features vectors as shown in Algorithm~\ref{alg2minmaxmean}.  Here, the inputs are the feature vectors (FC6 and  FC7) and the outputs are 3 merged features vectors based on the following equations:

\begin{eqnarray}\label{E:FFVmin}
 	FFVmin_i = \min{(FC6_i,  FC7_i)} 
\end{eqnarray}
\begin{eqnarray}\label{E:FFVmax}
 	FFVmax_i = \max{(FC6_i,  FC7_i)} 
\end{eqnarray}
\begin{eqnarray}\label{E:FFVmean}
FFVmean_i = \frac{FC6_i + FC7_i}{2} 
\end{eqnarray}

\begin{table}
\centering
	\caption{The number of features after the PCA applied on the feature vector resulting from merging FC6 and FC7}\label{table34}
	\begin{tabular}{l|l|llll}
	\hline
	&	Method	&	Face 	&	Gender 	&	Age		&	 Expression\\
		\hline
	Raw 		&	Min	&	4096	&	4096	&	4096	&	4096\\
	features	&	Max	&	4096		&	4096		&	4096		&	4096\\
 	\hline
	PCA 99\%	&	Min	&	734	&	734	&	734	&	734\\
	&	Max	&	745	&	745	&	745	&	745\\
	&	Mean	&	424	&	424	&	424	&	424\\
	\hline
	PCA 97\%	&	Min	&	356	&	356	&	356	&	356\\
	&	Max	&	354	&	364	&	364	&	364\\
	&	Mean	&	195	&	195	&	195	&	195\\
	\hline
	PCA 95\%	&	Min	&	219	&	219	&	219	&	219\\
	&	Max	&	224	&	224	&	224	&	224\\
	&	Mean	&	127	&	127	&	127	&	127\\
	\hline
	\end{tabular}	
\end{table}	

The dimensions of the resulting feature vector FFV remains the same after merging both of (FC6 and  FC7), therefore, we applied the PCA For dimensionality of each of the three merged vectors. Table~\ref{table34} shows the number of features for each classification tasks after applying the three percentages of PCA on the merged feature vectors.

\section{Experimental Results and Discussion}\label{sec:results}

We used 10-fold cross-validation to evaluate the proposed model. The advantage of this method over repeated random sub-sampling is that all observations are used for both training and testing. Our methods were evaluated using a set of machine learning classifiers; namely, k-nearest neighbors (kNN), random forest (RF), Na\"{i}ve Bayes (NB), BayesNet (BN), and artificial neural network (ANN); mainly for identifying persons and recognizing gender, age and facial expression (eye-smil) all from  veiled-face images. PCA is used for dimensionality reduction, and merging both FC6 and FC7 is used for obtaining more representative deep features.

\subsection{Results of Veiled-Face Recognition}\label{ssec:facerecog}

Initially, we evaluated the ability of a VGG19 to identify persons from VPI-New dataset by applying a number of classifiers on the both of the raw deep features vectors (FC6 and FC7) extracted by VGG19.

\begin{table}
\centering
	\caption{The accuracy of identifying persons from VPI-New dataset on raw deep features and after applying PCA with $99\%$, $97\%$, $95\%$ data variance, data in ($\%$).}\label{table41}
	\begin{tabular}{l|l|lllllll}
	\hline
	&	layer	&	1NN	&	3NN	&	5NN	&	RF	&	NB	&	BN	&	ANN\\
	\hline
	Raw 	&	FC6	&	\textbf{99.5238}	&	98.7619	&	98.4286	&	84.6667	&	98.0476	&	98.0476	& -\\
	features&		FC7	&	98.7143	&	97.1905	&	96.8571	&	83.2381	&	95.381	&	93.619	&	-\\
	\hline
	PCA 	&	FC6	&	98.7143	&	97.4762	&	96.1429	&	68.000	&	94.7143	&	96.619	&	\textbf{99.9524}\\
	99\%	&	FC7	&	99.4286	&	98.9524	&	98.7619	&	74.7619	&	97.7143	&	89.9048	&	99.8095\\
	\hline
	PCA 	&	FC6	&	99.619	&	99.4762	&	99.2857	&	79.5238	&	97.7143	&	96.619	&	\textbf{99.9048}\\
	97\%	&	FC7	&	99.2857	&	98.7143	&	98.6667	&	55.619	&	96.8571	&	97.7619	&	99.7619\\
	\hline	
	PCA 	&	FC6	&	99.7143	&	99.619	&	99.381	&	83.2381	&	98.381	&	96.619	&	\textbf{99.9524}\\
	95\%	&	FC7	&	99.3333	&	98.8571	&	98.7143	&	81.5238	&	97.5714	&	89.9048	&	99.4286\\
	\hline
\end{tabular}	
\end{table}

As can be seen from Table~\ref{table41} the deep features obtained from FC6 allow for higher accuracy of identifying persons than those obtained from FC7, where the highest accuracy  was $99.5238\%$ recorded by the 1NN classifier. Normally, the dimensionality of the deep features is high, that why we used the PCA, which significantly reduces the dimensions while preserving important information, and speeds up the identification process. Reducing the dimensionality allows for using other classifiers such as the ANN, which needs unacceptable training time on the high dimensional deep features. 
It is also interesting to note that the accuracy of the classifiers applied on the features vector extracted from FC6 showed a relative decrease (in general) when applying PCA. Unlike the FC7, which allows the classifiers to benefit form applying PCA in most cases. Similarly, the highest obtained accuracy is $99.9524\%$ recorded by the ANN on the smaller set of features, i.e. after applying the PCA preserving $95\%$ of the data variance. The reason behind this performance might be due to removing  a large number of poor (redundant) deep features. Because of the fluctuating performance of FC6 and FC7,  we opt for merging these deep features together to get a new feature vector based on both, using either of arithmetic mean, maximum or minimum of FC6 and FC7. 

\begin{table}
\centering
	\caption{The accuracy of identifying persons from VPI-New dataset after applying three merge methods and after applying PCA with $99\%$, $97\%$, $95\%$ data variance, data in ($\%$).}\label{table45}
	\begin{tabular}{l|l|lllllll}
	\hline
	&	Method	&	1NN	&	3NN	&	5NN	&	RF	&	NB	&	BN		&	ANN\\
	\hline
	Raw 		&	min		&	99.1905	&	98.5238	&	97.7143	&	82.7619	&	97.381	&	96.2857	&	-\\
	features	&	max		&	99.381	&	98.9048	&	98.3333	&	84.8571	&	97.2857	&	96.5238	&	-\\
			&	mean	&	\textbf{99.4286}	&	98.8571	&	98.3333	&	85.9524	&	97.8095	&	97.381	&	-\\
	\hline
	PCA 	&	min	&	91.3333	&	83.8571	&	78.4762	&	63.381	&	89.619	&	97.381	&	99.9048\\
	99\%		&	max	&	90.8095	&	81.7143	&	78.5238	&	60.4286	&	88.7143	&	96.7143	&	99.8095\\
		&	mean&	98.9524	&	97.7143	&	96.9048	&	69.8571	&	95.3333	&	97.0952	&	\textbf{99.9524}\\
	\hline
	PCA &	min	&	99.2381	&	98.5714	&	97.8095	&	78.4762	&	95.2857	&	97.381	&	\textbf{99.9524}\\
	97\%	&	max	&	99.1429	&	98.2381	&	97.9524	&	47.2381	&	92.5714	&	97.4762	&	99.8095\\
	&	mean&	99.619	&	99.5714	&	99.1905	&	79	&	98.2857	&	97.0952	&	99.9048\\
	\hline
	PCA 	&	min	&	99.5714	&	99.3333	&	99		&	82.4762	&	97.2857	&	97.381	&	99.9048\\
	95\%	&	max	&	99.619	&	99.5238	&	99.381	&	80.619	&	97.2381	&	96.7143	&	99.8571\\
	&	mean&	99.7143	&	99.619	&	99.2857	&	83.2381	&	98.5714	&	97.0952	&	\textbf{99.9524}\\
	\hline
	\end{tabular}	
\end{table}

We compared each merging method as shown in Table~\ref{table45}, the best way to do the merge is the arithmetic mean, where the highest identification accuracy was recorded by most of the classifiers used. However, the highest accuracy obtained by merging was $99.4286\%$ recorded by the 1NN classifier, which is slightly less than that recorded on the raw deep features by the same classifier. However, using PCA after merging, particularly, merging by the arithmetic mean, allowed for better results in general, reaching up to $99.9524\%$ in more cases than without merging.

\subsection{Results of Veiled-Face Gender Recognition}\label{ssec:genderrecog}

\begin{table}
\centering
	\caption{The accuracy of gender recognition from VPI-New dataset with raw features and after applying PCA with $99\%$, $97\%$, $95\%$ data variance, data in ($\%$).}\label{table410}
	\begin{tabular}{l|llllllll}
	\hline
	&	layer	&	1NN	&	3NN	&	5NN	&	RF	&	NB	&	BN	&	ANN\\
	\hline
	Raw 		&	FC6	&	\textbf{99.8095}	&	99.5714	&	99.4286	&	92.1905	&	76.5714	&	78.381	&	- \\
	features	&	FC7	&	99.5238	&	99.0952	&	98.9048	&	92.6667	&	70.5714	&	72.7143	&	- \\
	\hline
	PCA 		&	FC6	&	99.5238	&	98.9524	&	98.2381	&	80.381	&	80.2381	&	86.2857	&	99.2381\\
	99\%		&	FC7	&	\textbf{99.8095}	&	99.7143	&	99.4762	&	90.6667	&	91.7619	&	87.5238	&	94.8095\\
	\hline	
	PCA 		&	FC6	&	\textbf{99.9048}	&	99.9524	&	99.619	&	90.3333	&	92.6667	&	86.4286	&	97.8571\\
	97\% 	&	FC7	&	99.8095	&	99.7619	&	99.4286	&	93.5714	&	90.381	&	87.7143	&	96.7619\\
	\hline	
	PCA 		&	FC6	&	99.8571	&	\textbf{99.9048}	&	99.8095	&	93.3333	&	91.7619	&	86.4286	&	95.8571\\
	95\%		&	FC7	&	99.8571	&	99.5714	&	99.381	&	93.9048	&	90.1905	&	87.381	&	96.6667\\
	\hline
	\end{tabular}	
\end{table}

In this section we tested the ability of a VGG19 to gender recognition from VPI-New dataset. A number of classifiers have been applied on the raw feature vectors extracted by the VGG19. Table~\ref{table410} shows the accuracy of gender recognition from VPI-New database. The highest accuracy of gender recognition obtained was $99.8095\%$, which is recorded by the 1NN classifier. The results also suggest that the deep features obtained from FC6 are more representative than those obtained from FC7.

As shown in Table\ref{table410}, the highest accuracy of gender recognition obtained after applying PCA of $99\%$ is $99.8095\%$, which is recorded by the 1NN classifier, this is always expected since preserving more data variance allows for more information to feed the learning algorithm. However, what is more interesting is to get better results using less information, e.g. PCA preserving $97\%$ or even $95\%$ of data variance recorded higher classification accuracy.

\begin{table}
\centering
	\caption{The accuracy of gender recognition from VPI-New dataset after applying three merge methods with raw features and after applying PCA with $99\%$, $97\%$, $95\%$ data variance, data in ($\%$).}\label{table411}
	\begin{tabular}{l|l|lllllll}
	\hline
	&	Method	&	1NN	&	3NN	&	5NN	&	RF	&	NB	&	BN		&	ANN\\
	\hline
	Raw 		&	min		&	99.5238	&	99.1429	&	98.8571	&	94.8095	&	74.2857	&	74.381&	-\\
	features	&	max		&	\textbf{99.619}	&	99.4286	&	99.1905	&	92.381	&	75.4286	&	76.0476	&	-\\
		&	mean	&	99.5238		&	99.381	&	98.9524	&	93.0476	&	74.5238	&	76.3333	&	-\\
	\hline
	PCA 	&	min	&	96.0476	&	93.381	&	91.7619	&	78.1429	&	64.0476	&	87.1429	&	98.5714\\
	99\%	&	max	&	97.9048	&	94.7619	&	90.7143	&	77.0476	&	76.0476	&	88		&	99.1429\\
	&	mean&	\textbf{99.619}	&	99.1905	&	98.9048	&	79.7619	&	82.0952	&	86.9524	&	98.8095\\
	\hline
	PCA 	&	min	&	99.7143	&	99.6667	&	99.2381	&	84.5238	&	70.2381	&	88.2857	&	98.9571\\
	97\%	&	max	&	99.7143	&	99.6667	&	99.381	&	82.7619	&	90.7619	&	87.8095	&	99.1429\\
	&	mean&	\textbf{99.7619}	&	99.9048	&	\textbf{99.7619}	&	86.4286	&	92.2857	&	87.4286	&	97.9048\\
	\hline
	PCA 	&	min	&	99.7143	&	99.7143	&	99.7619	&	87.0952	&	82.2857	&	88.0476	&	98.1905\\
	95\%	&	max	&	99.8095	&	99.8571	&	99.7619	&	87.2857	&	93.0952	&	87.8095	&	98.1429\\
	&	mean&	99.8095	&	\textbf{99.9048}	&	99.8095	&	88.8095	&	92.2381	&	86.9048	&	95.9524\\
	\hline
	\end{tabular}	
\end{table}

Table~\ref{table411} shows the accuracy of gender recognition of each of the three feature vectors resulting from merging FC6 and FC7. As can be seen from this table, the merging methods produced no better classification. However, the accuracy of gender recognition from the merged deep features has increased after applying PCA, but it is still slightly less than the maximum accuracy obtained, except for the 3NN with (again) PCA $95\%$, which recorded the same ($99.9048\%$) using the mean merger.

\begin{table}
\centering
	\caption{The confusion matrix resulting from applying 3NN classifier after applying PCA95\%, and the total accuracy is 0.999048.}\label{table414}
	\begin{tabular}{l|ll}
	\hline
		-	&	Female	&	Male\\
		\hline
	Female	&	1539	&	1\\
	Male		&	1	&	559\\
	\hline
	\end{tabular}	
\end{table}

Table~\ref{table414} shows the structure of the confusion matrix where TP refers to true positive, FN refers to a false negative, FP refers to false positive and TN refers to true negative. The confusion matrix resulting from applying 3NN classifier after applying PCA by preserving $95\%$ of data variance. Note that there were two classes - Female, Male.

\subsection{Results of Veiled-Face Age Recognition}\label{ssec:agerecog}

\begin{table}
\centering
	\caption{The accuracy of age recognition from raw features and after applying PCA with $99\%$, $97\%$, $95\%$ data variance, data in ($\%$).}\label{table415}
	\begin{tabular}{l|l|lllllll}
	\hline
	&	layer	&	1NN	&	3NN	&	5NN	&	RF	&	NB	&	BN	&	ANN\\
	\hline
	Raw		&	FC6	&	99.9524	&	99.5714	&	99.2857	&	88.8095	&	89.8571	&	71.6667	&	-\\
	features	&	FC7	&	99.3333	&	99.0952	&	98.6667	&	89.1905	&	59.1429	&	63.5238	&	-\\
	\hline
	PCA &	FC6	&	98.8095	&	97.9048	&	95.3333	&	56.5714	&	79.9048	&	59.8571	&	99.4762\\
	99\%	&	FC7	&	99.5238	&	99.6667	&	99.381	&	79.7143	&	90.5714	&	83.7619	&	97\\
	\hline
	PCA 	&	FC6	&	99.8095	&	99.7143	&	99.5238	&	82.9048	&	92.1429	&	83.7143	&	99.7143\\
	97\%	&	FC7	&	99.8095	&	99.6667	&	99.5714	&	84.381	&	87.2857	&	83.9048	&	96.1905\\
	\hline
	PCA 	&	FC6	&	100		&	99.9524	&	99.8571	&	86.619	&	90.3333	&	83.7619	&	97.8524\\
	95\%	&	FC7	&	99.8095	&	99.6667	&	99.4286	&	88.1429	&	86.9048	&	83		&	96.4762\\
	\hline
	\end{tabular}	
\end{table}

We applied a number of classifiers  on the raw deep features extracted by the VGG19 for age recognition from VPI-New dataset. Table~\ref{table415} shows the accuracy of age recognition. The highest age recognition accuracy obtained was $99.9524\%$, which is recorded by the 1NN classifier on the FC6 deep features. As can be seen from the results in Table~\ref{table415}, the performance of the features obtained from layer FC6 for the purpose of age recognition, is more representative than that of layer FC7.

The most interesting note in Table~\ref{table415}, is the zero error rate, which is obtained by the 1NN classifier, after applying the PCA $95\%$ on the deep features obtained from FC6. What makes this result special is that the classifier used only $137$ features out of $4096$ features, which are obtained by the PCA after preserving $95\%$ of data variance. And this, surprisingly, allows for fast and perfect results. Such a perfect result might be due to the elimination of redundant and poor deep features.

\begin{table}
\centering
\caption{The accuracy of age recognition from VPI-New dataset after applying the three merging methods on the raw features and after applying PCA with $99\%$, $97\%$, $95\%$ data variance, data in ($\%$).}\label{table419}
\begin{tabular}{l|l|lllllll}
\hline
&	Method	&	1NN	&	3NN	&	5NN	&	RF	&	NB	&	BN		&	ANN\\
\hline
Raw 		&	min	&	99.8095	&	99.619	&	99.1905	&	89.8095	&	67.8095	&	69		&	-\\
features	&	max	&	99.7143	&	99.7143	&	99.3333	&	89.0476	&	66.9048	&	69		&	-\\
		&	mean&	\textbf{99.8571}	&	99.6667	&	99.1429	&	90.4286	&	68.8571	&	70.9048	&	-\\
\hline
PCA 		&	min	&	92.9524	&	86		&	80.5238	&	68		&	68.6667	&	84.1905	&	99.1429\\
99\%		&	max	&	93.9524	&	88.2381	&	83.1429	&	62.1429	&	68.619	&	82.5714	&	99.169\\
		&	mean&	99.4286	&	98.8571	&	97.619	&	73.381	&	83.2381	&	84.3333	&	99.6667\\
	\hline
PCA 		&	min	&	99.619	&	98.8095	&	98		&	79.7143	&	85.4286	&	84.1905	&	99.5714\\
97\%		&	max	&	99.4762	&	99		&	98.4762	&	76.9524	&	86.0952	&	82.5714	&	99.619\\
		&	mean&	99.8095	&	99.8571	&	99.5714	&	84.4762	&	91.5714	&	84.381	&	99.1905\\
	\hline	
PCA 		&	min	&	99.8095	&	99.7143	&	99.5714	&	86.1429	&	89.0952	&	84.1905	&	99.1429\\
95\%		&	max	&	99.8571	&	99.7619	&	99.5238	&	82.0476	&	91.8095	&	82.5238	&	99.0476\\
		&	mean&	\textbf{99.9524}	&	99.9524	&	99.7619	&	86.1429	&	89.8571	&	84.6667	&	97.7143\\
	\hline
\end{tabular}	
\end{table}

We further merged both FC6 and FC7, using the arithmetic mean, maximum and minimum for the purpose of age recognition. Table~\ref{table419} shows the accuracy of age recognition for each of the three feature vectors, which were resulted from the three merging processes. As can be seen from the same table, all of the merged features have outperformed the results of the raw deep features of FC7, as well as some of the results of the raw deep features of FC6. The highest age recognition result ($99.9524\%$) obtained after applying PCA $95\%$ on the merged features of the arithmetic mean, this high recognition result, again, fosters the ability of  merging the deep features, particularly, when using the arithmetic mean. 

\begin{table}
\centering
	\caption{The confusion matrix of the 1NN classifier  on PCA $95\%$ of the arithmetic mean merged features.}\label{table421}
	\begin{tabular}{l|llll}
	\hline
	-	&	children	&	youth	&	adults	&	elderly\\
	\hline
	Children	&	503	&	0		&	1	&	0\\
	Youth	&	0	&	1050		&	0	&	0\\
	Adults	&	0	&	0		&	462	&	0\\
	Elderly	&	0	&	0		&	0	&	84\\
	\hline
\end{tabular}	
\end{table}

Table~\ref{table421} shows the confusion matrix, which is resulted from applying 1NN classifier on the PCA $95\%$ of the arithmetic mean merged deep features, where we have four classes - Children, Youth, Adults and Elderly.

\subsection{Result of Veiled-Facial Expression Recognition}\label{ssec:expressionrecog}

\begin{table}
\centering
	\caption{The accuracy of facial expression (eye smile) recognition from VPI-New dataset on the raw deep features and after applying PCA with $99\%$, $97\%$, $95\%$ data variance, data in ($\%$).}\label{table422}
	\begin{tabular}{l|l|lllllll}
	\hline
	&	layer	&	1NN	&	3NN	&	5NN	&	RF	&	NB	&	BN	&	ANN\\
	\hline
	Raw 		&	FC6	&	\textbf{79.1905}	&	75.4762	&	71.8571	&	73.6667	&	61.7143	&	64.6667	&	-\\
	features	&	FC7	&	78.619	&	75.1429	&	71.8095	&	67.381	&	62.619	&	66.2857	&	-\\
	\hline
	PCA 	&	FC6	&	78.619	&	75.4762	&	71.8571	&	72.2857	&	71.8571	&	71.8571	&	78.6667\\
	99\%	&	FC7	&	78.5238	&	75.5238	&	71.381	&	72.0476	&	70.9524	&	70.9524	&	77.381\\
	PCA 	&	FC6	&	78.7619	&	76.1429	&	71.381	&	73.4286	&	77.1429	&	71.8571	&	\textbf{80}\\
	97\%	&	FC7	&	78.4762	&	75.2857	&	71.1905	&	73.2381	&	74.3333	&	70		&	77.0952\\
	PCA 	&	FC6	&	79.0952	&	75.2857	&	70.7619	&	73.7619	&	76.5238	&	71.9524	&	79.9524\\
	95\%	&	FC7	&	77.8095	&	74.4732	&	70.5714	&	73.5238	&	72.9524	&	70.5714	&	76.2381\\
	\hline
	\end{tabular}	
\end{table}

Table~\ref{table422} shows the accuracy of facial expression (eye-smile) recognition, since the eye-smile is the only facial expression recorded in the VPI-new database. As can be seen from the same table, the highest accuracy ($80\%$) is recorded by the ANN on FC6 after applying PCA, and in general, FC6 provides more distinctive deep features than that of FC7.

\begin{table}
\centering
	\caption{The accuracy of facial expression (eye smile) recognition from VPI-New dataset after merging and applying PCA by $99\%$, $97\%$, $95\%$ data variance, data in ($\%$).}\label{table424}
	\begin{tabular}{l|l|lllllll}
	\hline
	&	Method	&	1NN	&	3NN	&	5NN	&	RF	&	NB	&	BN		&	ANN\\
	\hline
	Merge 	&	min	&	79.381	&	75.2857	&	72.2857	&	73.7619	&	61.2857	&	63.2381	&	-\\
	features	&	max	&	79.0476	&	75.4762	&	71.619	&	73.3333	&	65		&	66.5714	&	-\\
		&	mean&	\textbf{79.5714}	&	75.5714	&	72.2381	&	74.381	&	61.5238	&	64.3333	&	-\\
	\hline
	PCA 	&	min&		78.3333	&	74.8095	&	73.2857	&	70.3333	&	63.7619	&	70.8095	&	76.285\\
	99\%	&	max&	79.381	&	75.4762	&	72.0476	&	70.5714	&	64.3333	&	72.3333	&	75.3333\\
	&	mean&	78.9048	&	75.5714	&	72		&	61.619	&	71.7143	&	70.4762	&	77.5238\\
	\hline
	PCA 		&	min	&	78.8095	&	76.5714	&	71.6667	&	71.7143	&	71.0476	&	70.6667	&	79\\
	97\%		&	max	&	79.0476	&	76.7143	&	71.619	&	72.0952	&	68.6667	&	72.4286	&	78.9048\\
		&	mean&	78.619	&	76.2857	&	71.381	&	72.8571	&	75.7619	&	70.6667	&	78.9048\\
		\hline
	PCA 		&	min	&	78.7619	&	76.2857	&	71.6667	&	73.1429	&	75.1429	&	69.6667	&	\textbf{80.8571}\\
	95\%		&	max	&	79.7143	&	75.9048	&	71.619	&	72.5714	&	72		&	72.5714	&	80.333\\
		&	mean&	78.9524	&	75.381	&	71.6667	&	72.9048	&	75.6667	&	70.7143	&	79.2381\\
	\hline
	\end{tabular}	
\end{table}

Similar to the previous problems, we merged both FC6 and FC7 deep features, using the aforementioned merging methods. Table~\ref{table424} shows the accuracy of the facial expression (eye smile) recognition after each merging method. As expected, and similar to the previous problems, we can easily see  that the eye smile recognition has slightly improved  when using the merged deep features, with no significant preference to any merging method.

\begin{table}
\centering
	\caption{The confusion matrix of eye smile recognition by the ANN classifier of PCA $95\%$ on the minimum merged deep features.}\label{table426}
	\begin{tabular}{l|ll}
	\hline
		-	&	Smile	&	Normal\\
		\hline
	Smile		&	1313		&	187\\
	Normal		&	215	&	385\\
	\hline
	\end{tabular}	
\end{table}

Table~\ref{table426} shows the confusion matrix, which is resulted from applying the ANN classifier on the PCA $95\%$ of minimum merged deep features, where we have two classes - Smile, Normal.

Most of the previous results show that the use of PCA has improved the identification/recognition tasks. This is also supported by~\cite{tar2020,gan2015,linge2014}, whose experiments in face recognition have shown improvement in the performance of classification using PCA and neural network. Our results also show that the use of the deep features in general, and those obtained from FC6, in  particular, allowed for better recognition/identification. This is supported by the results of~\cite{kataoka2015}, whose results have shown that the fully connected layers tend to perform better for object recognition tasks.

\begin{table}
\centering
	\caption{Results of person identification, gender recognition, smile recognition and gender recognition of mean-merged deep features with PCA preserving $99\%$ of data.}\label{table4261}
	\begin{tabular} {lllllll}
	\cline{2-7}
	\multirow{2}{*}{}     & \multicolumn{3}{c}{1NN}         & \multicolumn{3}{c}{ANN}         \\ \cline{2-7} 
                      & F-Measure & ROC Area & PRC Area & F-Measure & ROC Area & PRC Area \\ \hline
	Identify Persons      & 0.99      & 0.995    & 0.981    & 0.999     & 1.000    & 1.000    \\ \hline
	Gender recognition    & 0.996     & 0.996    & 0.995    & 0.988     & 0.998    & 0.997    \\ \hline
	Smile recognition & 0.787     & 0.733    & 0.733    & 0.758     & 0.766    & 0.791    \\ \hline
	Age recognition       & 0.994     & 0.996    & 0.990    & 0.996     & 1.000    & 1.000    \\ \hline
	\end{tabular}
\end{table}

Sometimes, the accuracy measures is not enough to show the real performance of a machine learning method~\cite{olanrewaju2017enhancement}; particularly when applied to an imbalanced dataset~\cite{ghatasheh2020cost, tarawneh2020smotefuna, abu2019effects}. Although our dataset of the deep features for persons identification is balanced; having $14$ images for each subject/class, we have the other three datasets, which contain the deep features for gender, age, or smile recognition is not fairly balanced as can be seen from the confusion matrices in Tables~\ref{table414},~\ref{table421} and~\ref{table426} respectively. Therefore, we opt for more metrics, namely, F-Measure, ROC Area, and PRC Area. However, instead of repeating all the previous results, we opt for the results of deep features merged using the mean method, as being the best performer, and we used the PCA preserving $99\%$ of the data for the same reason. Table~\ref{table4261} shows the results with these metrics. As can be seen from Table~\ref{table4261}, and similar to the previous accuracy results, the high values of F-Measure, ROC Area, and PRC Area indicate the robustness of the extracted PCA features from the deep features obtained by mean-merging FC6 and FC7.

\subsection{Results comparison to some related work}\label{ssec:resultscompar}

\begin{table}
\centering
	\caption{The accuracy of identifying persons and gender recognition from VPI-Old dataset ~\cite{hassanat2017} using deep features, data in (\%). }\label{table427}
	\begin{tabular}{@{}l|l|llllll}
	\hline
	&	layer	&	1NN	&	3NN	&	5NN	&	RF	&	NB	&	BN	\\
	\hline
	Identify 	&	FC6	&	99.6667	&	99.5		&	99.4167	&	97.3333	&	94		&	94.4167\\
	persons	&	FC7	&	99.4167	&	99.4167	&	99.5		&	97.25	&	87.5833	&	88.25\\
	\hline
	Gender 	&	FC6	&	99.6667	&	99.5		&	99.4167	&	97.3333	&	94		&	94.4167\\
	recognition&	FC7	&	99.4167	&	99.4167	&	99.5		&	97.25	&	87.5833	&	88.25\\
	\hline
	\end{tabular}	
\end{table}

To the best of our knowledge, no methods have been proposed to identify persons and gender recognition from veiled-face images, except for ~\cite{hassanat2017}, who used hand-craft features on the VPI-Old database. Their highest identification rate was $97.55\%$, and $99.41\%$ for gender recognition. Both VPI-Old and new databases are different, therefore, and for the purpose of a valid comparison, we extracted the deep features using the same VGG19.As can bee seen from Table~\ref{table427}, the use of the deep features has improved the identification/recognition rates significantly.

\begin{table}
\centering
	\caption{Comparison of our proposed approach to other methods.}\label{table428}
	\begin{tabular}{l|llll}
	\hline
	Methods	&	Identify 	&	Gender 		&	Age 			&	facial expression \\
	&	persons	&	recognition	&	recognition	&	recognition\\
	\hline
	Our proposed &	99.95\%	&	99.9\%	&	100\%	&	80.9\%\\
	\cite{hassanat2017}  
	&	97.22\%	&	99.41\%	&	-	&	-\\
	\cite{teo2007} 	
	&	95.12\%	&	-	&	-	&	-\\
	\cite{hu2013}
	&	98\%	-	&	-	&	-\\
	\cite{liew2016}
	&	-	&	99.38\%	&	-	&	-\\
	\cite{sun2015} 
 	&	99.53\%	&	-	&	-	&	-\\
	\cite{lian2016} 
 	&	-	&	94.08\%	&	-	&	-\\
	\cite{yu2015}
 	&		-	&	-	&	-	&	61.29\%\\
	 
 	\cite{rothe2018}
 	&		-	&	-	&	 96.6\%	&	-\\
 	 
 	\cite{dehshibi2010new}
 	&		-	&	-	&	 86.64\%	&	-\\
	\hline
	\end{tabular}	
\end{table}

Although it is not a good practice to compare the proposed approach to other methods that used other databases, e.g. Table~\ref{table428}, at least, such a comparison gives a hint about the goodness of the deep features, when used for our problems. Knowing that the deep CNNs have recently achieved excellent performance in a wide range of image classification tasks~\cite{yu2015}, unlike traditional machine learning methods, where features are hand-crafted~\cite{mollahosseini2016}.

\section{Conclusions}\label{sec:conc}

In this paper, we proposed a new approach based on deep features to identify persons, and recognize genders, ages, and a facial expression (eye smile), all from veiled-faces image database, which was created for the purpose of this study. The proposed approach is basically based on extracting deep features from both FC6 and FC7 layers of the VGG19, merging these features and then reducing the dimensionality using PCA. We investigated a number of merging methods with different classifiers. 

Our results obtained on an in-house veiled-faces image database show that it is possible to identify persons, and recognize genders, ages, and a facial expression (eye smile), all from veiled-faces, the use of deep features is a good choice for our tasks, and the PCA of the deep features extracts more distinctive features for better identification/recognition rates in most cases, with a smaller number of features, and hence allows for faster learning/testing. In addition, we found that the 1NN and ANN classifiers (in general) were the best classifiers to be used for our tasks in terms of accuracy. We also found that merging deep features (FC6 and FC7), particularly, when using the arithmetic mean, provides better identification/recognition results in most cases. 

The limitation of this work is linked to the in-house image database used, since the images of the frontal veiled-faces where taken from a short distance, in an office environment, we used this for simplicity, so as not to be forced to use face detection, and handling complex backgrounds, etc. more than about the missing information posed by the usage of veils. A more realistic veiled-face image database (in the wild) is needed to further prove the feasibility of the proposed approach and to be used in practice. Our future will focus on collecting such a complex database, in addition to investigating more deep learning methods that can handle complex backgrounds robustly.  

\section*{Acknowledgment}
The third author (A. S. T.) would like to acknowledge Tempus Public Foundation for sponsoring his Ph.D. program. His work is also supported by the Hungarian Government and co-financed by the European Social Fund, under the project EFOP-3.6.3-VEKOP-16-2017-00001 (Talent Management in Autonomous Vehicle Control Technologies.
\bibliographystyle{apalike}
\bibliography{references.bib}

\end{document}